\documentclass[conference]{IEEEtran}
\IEEEoverridecommandlockouts
\usepackage{cite}
\usepackage{amsmath,amssymb,amsfonts}
\usepackage{algorithmic}
\usepackage{graphicx}
\usepackage{textcomp}
\usepackage{xcolor}

\usepackage{booktabs}
\usepackage{multirow}
\usepackage{authblk}
\usepackage{url}

\def\BibTeX{{\rm B\kern-.05em{\sc i\kern-.025em b}\kern-.08em
    T\kern-.1667em\lower.7ex\hbox{E}\kern-.125emX}}
\begin{document}

\title{AdaForensics: Learning A Characteristic-aware Adaptive Deepfake Detector
\thanks{* Corresponding author.}
}

\author[1]{Xiaoke Yang}
\author[2]{Haixu Song}
\author[2]{Xiangyu Lu}
\author[1]{Shao-Lun Huang}
\author[2,*]{Yueqi Duan}

\affil[1]{Tsinghua Shenzhen International Graduate School, Tsinghua University, Shenzhen, China}
\affil[2]{Department of Electronic Engineering, Tsinghua University, Beijing, China}
\affil[ ]{\texttt{\{yangxk22, shx22, luxiangy21\}@mails.tsinghua.edu.cn}}
\affil[ ]{\texttt{shaolun.huang@sz.tsinghua.edu.cn;duanyueqi@tsinghua.edu.cn}}

\maketitle

\begin{abstract}
In this paper, we propose a characteristic-aware adaptive network named AdaForensics for deepfake detection. Most existing methods learn a fixed network to detect deepfakes based on carefully-designed network architectures. However, these methods employ the same deepfake detector for all the images despite of various facial characteristic, which fail to provide customized forgery detection for different individuals. To address this, our AdaForensics simultaneously learns characteristic-agnostic and characteristic-specific embeddings, where the detector dynamically adapts to varying faces with our designed hypernetwork on the fly. More specifically, our AdaForensics not only explores the shareable abstractions from various deepfake images, but also adapts the detector to the given characteristic at test time. To achieve this, we propose a two-branch HyperNetwork to learn an adaptive deepfake detector, which automatically adjusts the parameters based on characteristic of the input. Extensive experiments on widely-used datasets including FaceForensics, Celeb-DF and DFDC demonstrate our AdaForensics outperforms the state-of-the-art works.
\end{abstract}


\section{Introduction}
\label{sec:intro}

In recent years, the rapid development of generative models \cite{kingma2013auto} has made it possible to make diverse modifications to face images easily and effectively. These models are capable of generating highly realistic media that is indistinguishable from real images to the human eye. While deepfake technology has great potential for positive applications, unscrupulous individuals create pornographic movies, fake news, and political rumors through deepfake, causing great negative effects. Therefore, to minimize the malicious misuse of deepfakes, effective deepfake detection methods are needed to help people determine the credibility of media.

So far, various carefully-designed detectors have been developed for deepfake detection. Conventional deepfake detectors attempt to develop optimal CNN architectures that primarily rely on local regions, blending boundaries, and global textures \cite{li2020face, chai2020makes}. In addition, some works \cite{liu2021spatial, luo2021generalizing, qian2020thinking} exploit the frequency differences in deepfakes to enhance performance. Some other effective approaches focus on specific representations, such as neuron behaviors, landmark geometric features, and attentional networks \cite{yan2023ucf}.

\begin{figure}
\label{teaser}
\centering
\includegraphics[width=0.9\linewidth]{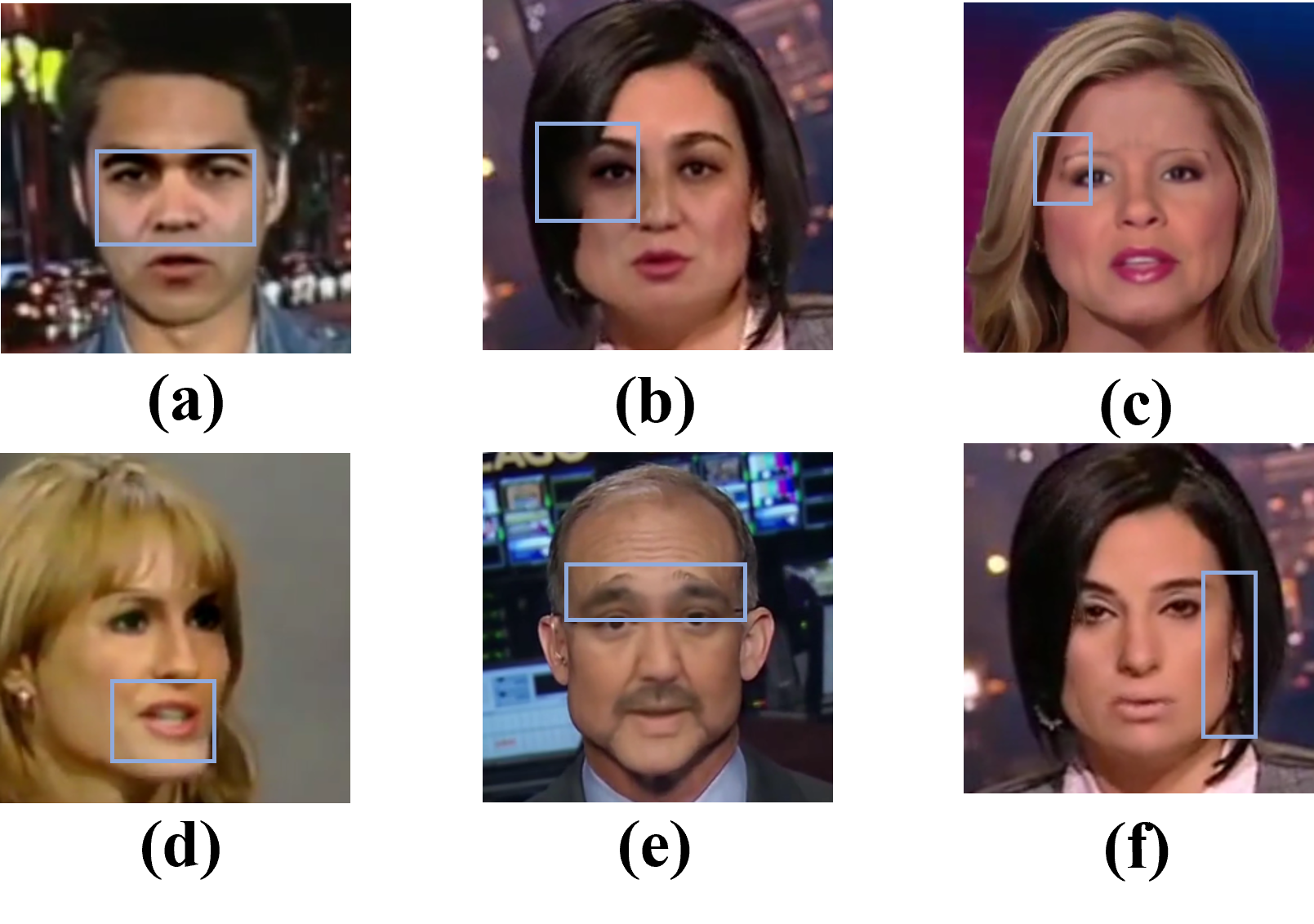}
\caption{There exists several forgery artifacts generated by Deepfake of FaceForensics++ \cite{roessler2019faceforensicspp} including: (a) different skin color, (b) missing hair, (c) elongated eyebrows, (d) slightly misaligned mouths, and (e) double eyebrows. Although (f) manipulated by Face2Face possesses the same original face with (b), it shows a noticeable gap in the facial edge.}
\end{figure}

In deepfake detection, the diversity of characteristics leads to a variety of forgery artifacts, challenging the ability of fixed networks to capture these significant differences. 
For example, as illustrated in Fig. \ref{teaser}, images (a)-(e) manipulated by the same method show different forgery artifacts influenced by the unique characteristics of each individual. Additionally, the manipulation methods themselves impact the characteristics of deepfakes even with the same original face, as observed in Fig. \ref{teaser}(f). 
Fixed detectors struggle to detect different forgery artifacts caused by various characteristics. This raises a natural question: can we develop a dynamic deepfake detector that is customized to these characteristic changes?

\begin{figure*}
    \centering
    \includegraphics[width=\linewidth]{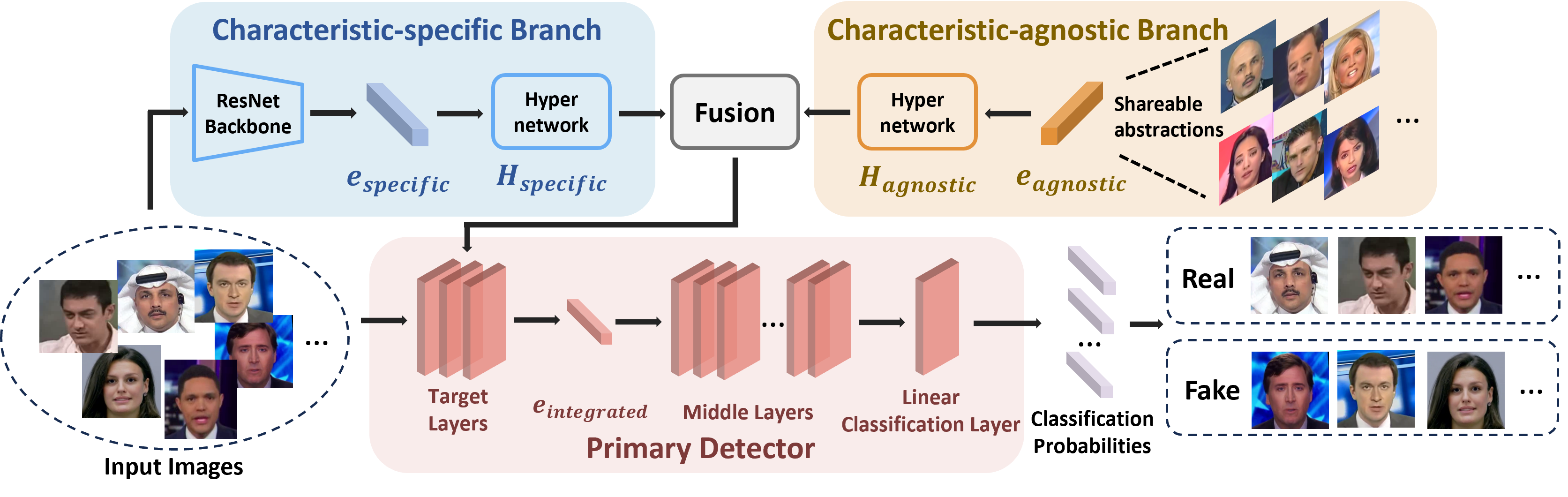}
    \caption{ Overview of our AdaForensics framework, which consists of three key components: the characteristic-specific branch, the characteristic-agnostic branch, and the primary detector. The characteristic-specific branch employs a ResNet backbone to extract characteristic information from the input image, which is then processed by a hypernetwork that generates a corresponding weight. Concurrently, the characteristic-agnostic branch captures shareable abstractions from a variety of deepfakes and generates another weight. Then the weights are fused and served as the target layer's weight, ensure the integration of both characteristic-specific and characteristic-agnostic information.  }
    \label{pipeline}
\end{figure*}

To address the aforementioned issues, we introduce AdaForensics, a characteristic-aware adaptive deepfake detector. 
Unlike existing methods that rely on static networks, we design our AdaForensics  to learn and integrate characteristic-conditioned information that is unique to each individual.
This personalized approach allows the integration of diverse, individual-specific knowledge into the network parameters, enabling our deepfake detector to dynamically adapt to the distinct characteristics of different input images. 
Specifically, the characteristic-conditioned knowledge can be factorized into two levels: characteristic-agnostic and characteristic-specific information. 
For characteristic-agnostic information, we maintain a learnable embedding that is consumed by a hypernetwork and continuously updated as input faces are parsed during training. 
This shareable information abstracts the attributes of training faces and can be exploited by the detector at test time. 
For characteristic-specific information, it customizes the detector for each specific face, ensuring effective detection of deepfakes in various facial images, even when they exhibit different forgery artifacts.
Our research shows that characteristic-conditioned knowledge provide prior to eliminate the uncertainty of various forgery artifacts across different deepfakes in detection. We evaluate our method with both intra-dataset test and cross-datase test on widely-used datasets, demonstrating that our framework outperform the current state-of-the-art methods.

\section{Related Work}

\textbf{Deepfake Detection.} Given the rapid rise in the prevalence and accessibility of deepfakes, researchers have been motivated to develop sophisticated methodologies for their detection. Early deepfake detection methods, such as MesoNet\cite{afchar2018mesonet} and CNN-Aug\cite{wang2020cnn}, primarily concentrated on identifying spatial artifacts present in manipulated images. These methods aimed to construct optimal convolutional neural network (CNN) detectors to discern between real and fake images effectively. Subsequent research has focused on detecting specific types of artifacts. For instance, methods like SPSL\cite{liu2021spatial}, SRM\cite{luo2021generalizing}, and F3Net\cite{qian2020thinking} have examined the frequency differences between real faces and deepfakes, identifying subtle cues such as compression artifacts that are often overlooked.

Moreover, Face X-ray\cite{li2020face} introduces the concept of a blending region in forged images, assuming its presence to improve detection accuracy. On the other hand, SBI\cite{shiohara2022detecting} delves into the analysis of blending artifacts and enhances detectors through specific data augmentation techniques designed to expose these subtle inconsistencies. FWA\cite{li2018exposing} focuses on maintaining the consistency between the inner and outer face regions, which is critical for detecting anomalies. Similarly, ICT\cite{dong2022protecting} addresses the consistency of these regions and also leverages freely available videos on the internet to bolster the training data, thereby improving detection robustness. DeepFakeFace\cite{song2023robustness} demonstrates the importance of data augmentation by applying previous methods to diffusion-generated datasets.

Additionally, RECCE\cite{cao2022end}, a method that exclusively reconstructs genuine faces, allows the model to learn compact representations that effectively distinguish real images from fake ones. In recent advancements, UCF\cite{yan2023ucf} has been dedicated to the challenging task of distinguishing features pertinent to forgery detection from those that are not. This approach aims to extract valuable information specific to deepfakes, enhancing the accuracy of detection systems.

Despite the considerable performance of these approaches, they are limited by the rigidity of their fixed network architectures and the lack of sample-specific knowledge. This inflexibility hinders their ability to adapt detection strategies based on the unique characteristics of each input, posing a significant challenge in the dynamic landscape of deepfake detection.

\textbf{HyperNetworks in Deep Neural Networks.} HyperNetworks \cite{ha2016hypernetworks} is an approach that employs a network, also referred to as a hypernetwork, conditioned on a specific input embedding to generate weights for the target network (also known as the primary network). The approach has been extended to various applications, including image recognition \cite{ha2016hypernetworks}, semantic segmentation \cite{nirkin2021hyperseg}, and natural language modeling \cite{ha2016hypernetworks}. 
Further research is required to develop a more relevant framework for face reenactment. 
HyperReenact \cite{bounareli2023hyperreenact} presents a novel approach that leverages the adaptive nature of hypernetworks to generate realistic talking head images, demonstrating remarkable robustness even under extreme head pose variations.
Furthermore, MH \cite{zhao2020meta} illustrates the efficacy of HyperNetworks for expeditious adaptation and demonstrates that HyperNetworks can learn to model the shared structure underlying a family of tasks. 
In addition, HyperDet3d\cite{zheng2022hyperdet3d}, HyperFormer\cite{zheng2023learning} and AdaAc\cite{zhang2023hoi} apply HyperNetworks to separate condition-specific knowledge and condition-agnostic knowledge, resulting in enhanced performance.
Building upon previous works, our AdaForensics constructs characteristic-specific and characteristic-agnostic embeddings for deepfake detection with excellent generalizability, and is, to our knowledge, the first to incorporate hypernetworks in this task.

\section{Methodology}
Due to the fact that deepfakes vary in forgery artifacts, we consider deepfake detection from a characteristic-conditioned point of view. We design our deepfake detector AdaForensics as a two-branch HyperNetwork \cite{ha2016hypernetworks} to dynamically detect deepfakes for different images. Details of our characteristic-condition branches are provided in Sec. \ref{ss:twobranch}, and the primary network is described in Sec. \ref{ss:primary}. In Sec \ref{ss:implem}, we provide full implementation details of our method.

\subsection{Overview}
Fig. \ref{pipeline} illustrates the architecture of AdaForensics, which contains three components: the characteristic-agnostic branch, the characteristic-specific branch, and the primary detector. 
Given an input image $I$ with label $l$, the characteristic-specific branch extracts characteristic information using a facial feature extractor $E$ to generate the embedding $\mathbf{e}_{\text{specific}}$.
This embedding is then consumed by the characteristic-specific hypernetwork $H{\text{specific}}$ to produce the corresponding weight $\mathbf{W}_{\text{specific}}$.
Simultaneously, the learnable characteristic-agnostic embedding $\mathbf{e}_{\text{agnostic}}$ is processed by characteristic-agnostic hypernetwork $H_{\text{agnostic}}$ to yield $\mathbf{W}_{\text{agnostic}}$.
Both weights are then fused to create our adaptive weight $\mathbf{W}$, which is applied to the target layer in the primary detector for adaptive deepfake detection of the input image $I$.

\subsection{Characteristic-Condition Branches}
\label{ss:twobranch}

To detect deepfakes with characteristic-conditioned knowledge, we design a two-branch HyperNetwork. The first branch is the characteristic-agnostic branch, produces weight $\mathbf{W}_{\text{agnostic}}$. These are derived from a shareable abstraction embedding $\mathbf{e}_{\text{agnostic}}$, which is common across all deepfake detections. The second branch is characteristic-specific branch, generates weight $\mathbf{W}_{\text{specific}}$. This come from a facial feature embedding $\mathbf{e}_{\text{specific}}$ extracted by the facial feature extractor $E$.

\noindent \textbf{Characteristic-agnostic Branch.} The purpose of this branch is to generate $\mathbf{W}_{\text{agnostic}} \in \mathbb{R}^{C_{out}\times C_{in}}$ for primary detector, where $C_{in}$ and $C_{out}$ present the input and output dimension of target convolution layer. For characteristic-agnostic knowledge, we maintain a characteristic-agnostic embedding vector $\mathbf{e}_{\text{agnostic}}$, which is consumed by a characteristic-agnostic hypernetwork $H_{\text{agnostic}}$. Therefore, the characteristic-agnostic branch can be written as below:
\begin{equation}
    \mathbf{W}_{\text{agnostic}}=H_{\text{agnostic}}(\mathbf{e}_{\text{agnostic}}).
\end{equation}
While characteristic-agnostic knowledge is learned by our characteristic-agnostic branch and injected into the target layer, it can be combined with the other pretrained layers of our primary detector, which is enabled by the effectiveness of the HyperNetwork for quick adaptation \cite{ha2016hypernetworks, zhao2020meta}.

\noindent \textbf{Characteristic-specific Branch.} For the characteristic-specific branch, we also learn an embedding vector $\mathbf{e}_{\text{specific}}$ similar to $\mathbf{e}_{\text{agnostic}}$. 
The difference between them is that $\mathbf{e}_{\text{specific}}$ is adapted to the input face images, which allows the characteristic-specific hypernetwork $H_{\text{specific}}$ to employ the input face $I$ as a characteristic-specific query. 
To fully extract facial features from the input images, we exploit the feature extraction ResNet backbone of the well-known face identification method AdaFace \cite{kim2022adaface} and fine-tune its weights during training. 
This branch is represented by the following equations:
\begin{equation}
    \begin{cases}
        \mathbf{e}_{\text{specific}}=E(I) \\
        \mathbf{W}_{\text{specific}}=H_{\text{specific}}(\mathbf{e}_{\text{specific}})        
    \end{cases}.
\end{equation}
The dimensions of $\mathbf{W}_{\text{specific}}$ and $\mathbf{W}_{\text{agnostic}}$ are identical, which simplifies the implementation process and enhances the effectiveness of their fusion. The fusion of $ W_{\text{agnostic}}$ and $\mathbf{W}_{\text{specific}}$ is shown below:
\begin{equation}
    \mathbf{W} = \mathbf{W}_{\text{agnostic}}\odot \mathbf{W}_{\text{specific}},
\end{equation}
where $\mathbf{W} \in \mathbb{R}^{C_{out}\times C_{in}}$, has the same dimensions as the target layer.

\begin{table*}[h]
\centering
\caption{Intra-dataset AUC comparisons with top-tier methods trained on FF++c23, highlighting the best results in bold.}
\label{intra-test}
\resizebox{\linewidth}{!}{%
\begin{tabular}{ccccccccc}
\toprule
Detector                         & FF++c23       & FF++c40       & FF-DF           & FF-F2F          & FF-FS           & FF-NT           & Avg.                        & Top3       \\ \midrule
\multicolumn{1}{c|}{Meso4 \cite{afchar2018mesonet}}       & 0.6077          & 0.5920           & 0.6771          & 0.6170           & 0.5946          & 0.5701          & \multicolumn{1}{c|}{0.6097} & 0          \\
\multicolumn{1}{c|}{MesoIncepion4 \cite{afchar2018mesonet}}   & 0.7583          & 0.7278          & 0.8542          & 0.8087          & 0.7421          & 0.6517          & \multicolumn{1}{c|}{0.7571} & 0          \\
\multicolumn{1}{c|}{FWA \cite{li2018exposing}}         & 0.8765          & 0.7357          & 0.9210           & 0.9000             & 0.8843          & 0.8120           & \multicolumn{1}{c|}{0.8549} & 0          \\
\multicolumn{1}{c|}{EfficientB4 \cite{tan2019efficientnet}} & 0.9567          & 0.8150           & 0.9757          & 0.9758          & 0.9797          & 0.9308          & \multicolumn{1}{c|}{0.9389} & 0          \\
\multicolumn{1}{c|}{Capsule \cite{nguyen2019capsule}}     & 0.8421          & 0.7040           & 0.8669          & 0.8634          & 0.8734          & 0.7804          & \multicolumn{1}{c|}{0.8217} & 0          \\
\multicolumn{1}{c|}{Xception \cite{chollet2017xception}}    & 0.9637          & 0.8261          & 0.9799          & 0.9785          & 0.9833          & 0.9385          & \multicolumn{1}{c|}{0.9450}  & 4          \\
\multicolumn{1}{c|}{CNN-Aug \cite{wang2020cnn}}     & 0.8493          & 0.7846          & 0.9048          & 0.8788          & 0.9026          & 0.7313          & \multicolumn{1}{c|}{0.8419} & 0          \\
\multicolumn{1}{c|}{F3Net \cite{qian2020thinking}}       & 0.9635          & 0.8271          & 0.9793          & 0.9796          & 0.9844          & 0.9354          & \multicolumn{1}{c|}{0.9449} & 1          \\
\multicolumn{1}{c|}{FFD \cite{dang2020detection}}         & 0.9624          & 0.8237          & 0.9803          & 0.9784          & 0.9853          & 0.9306          & \multicolumn{1}{c|}{0.9434} & 1          \\
\multicolumn{1}{c|}{Face X-ray \cite{li2020face}}  & 0.9592          & 0.7925          & 0.9794          & 0.9872          & 0.9871          & 0.9290           & \multicolumn{1}{c|}{0.9391} & 3          \\
\multicolumn{1}{c|}{SPSL \cite{liu2021spatial}}        & 0.9610           & 0.8174          & 0.9781          & 0.9754          & 0.9829          & 0.9299          & \multicolumn{1}{c|}{0.9408} & 0          \\
\multicolumn{1}{c|}{SRM \cite{luo2021generalizing}}         & 0.9576          & 0.8114          & 0.9733          & 0.9696          & 0.9740           & 0.9295          & \multicolumn{1}{c|}{0.9359} & 0          \\
\multicolumn{1}{c|}{CORE \cite{ni2022core}}        & 0.9638          & 0.8194          & 0.9787          & 0.9803          & 0.9823          & 0.9339          & \multicolumn{1}{c|}{0.9431} & 2          \\
\multicolumn{1}{c|}{Recce \cite{cao2022end}}       & 0.9621          & 0.8190           & 0.9797          & 0.9779          & 0.9785          & 0.9357          & \multicolumn{1}{c|}{0.9422} & 1          \\
\multicolumn{1}{c|}{UCF \cite{yan2023ucf}}         & 0.9705          & \textbf{0.8399} & 0.9883          & 0.9840           & 0.9896          & 0.9441          & \multicolumn{1}{c|}{0.9527} & \textbf{7}          \\ \midrule
Ours                             & \textbf{0.9889} & 0.8263          & \textbf{0.9917} & \textbf{0.9903} & \textbf{0.9902} & \textbf{0.9621} & \textbf{0.9583}             & \textbf{7} \\ \bottomrule
\end{tabular}%
}
\end{table*}

\subsection{Primary Network}
\label{ss:primary}
Since HyperNetworks \cite{ha2016hypernetworks} commonly apply their generated weights to convolution layers or linear layers, we carefully designed our primary network based on Xception \cite{chollet2017xception}, a classic and efficient convolutional neural network. Xception \cite{chollet2017xception} consists of two connected entry convolution layers, a linear classification layer at the exit, and a repeating residual block in the middle. In order to make our model fully extract characteristic-conditioned information, we choose the first convolutional layer, as our weights injection layer, which can be specified from Fig. \ref{pipeline}. Experiments show that our framework can perform against the state-of-the-art works. Assuming that the weight-injected layer is denoted as $F_{\text{injected}}$, and subsequent layers of the primary network can be represented by $N_{\text{subsequent}}$, the primary network can be illustrated as follows:
\begin{equation}
\begin{cases}
    \mathbf{e}_{\text{integrated}} = F_{\text{injected}}(\mathbf{W},I) \\
    prob = N_{subsequent}(\mathbf{e}_{\text{integrated}})
\end{cases},
\end{equation}
where $\mathbf{e}_{\text{integrated}} \in \mathbb{R}^{C_{out}}$ is the output of the weight-injected layer, and $prob$ is a $\mathbb{R}^{2}$ probability representing the probability of the image classification result. Finally, we compare the predicted result with the true label $l$ to calculate our loss:
\begin{equation}
    L=\sum_{n}||\text{argmax}_{i}\ prob(i)-l||,
\end{equation}
where $\text{argmax}_{i}\ prob(i)$ returns the index of the maximum element, essentially the classification result of the network, while $n$ represents the batch size.

\subsection{Implenmentation Details} 
\label{ss:implem}
All our experiments are implemented with single NVIDIA RTX 3090. We modify Xception \cite{chollet2017xception} and AdaFace \cite{kim2022adaface} as the backbone for our method. The Xception \cite{chollet2017xception} backbone is initialized with pretrained weights from ImageNet\cite{deng2009imagenet} dataset, and AdaFace\cite{kim2022adaface} is based on ResNet50\cite{he2016deep} and initialized with MS1MV2\cite{deng2019arcface}. Face extraction and alignment are performed using Dlib\cite{king2009dlib}. The aligned faces are resized to 256 × 256 for the Xception\cite{chollet2017xception} backbone and 112 × 112 for the AdaFace\cite{kim2022adaface} backbone. The dimension of both hypernetwork embedding vectors is set to 256. We use the Adam for optimization with the learning rate of 0.0002 and the weight decay of 0.0005, and the batch size is fixed to 20. Additionally, we use a step scheduler with a step size of 10000 iterations and $\gamma=0.7$. We also apply some widely used data augmentations such as horizontal flip, rotation, image compression, and brightness contrast randomly. We consistently sample 32 frames from each video in the datasets.

\section{Experiments}
In this section, we evaluate the performance of our AdaForensics framework on popular benchmark datasets\cite{roessler2019faceforensicspp, li2020celeb, dolhansky2019deepfake, yang2019exposing, dolhansky2020deepfake}. We conduct intra-dataset evaluation and cross-dataset evaluation of our proposed method by benchmarking it against several state-of-the-art deepfake detection approaches. In addition, we perform ablation studies to evaluate the effectiveness of our proposed characteristic-conditioned knowledge. All tests demonstrate the superiority of our proposed method.

\subsection{Settings}
\noindent \textbf{Datasets.} We conduct extensive experiments on Faceforensics++ (FF++) \cite{roessler2019faceforensicspp}, Celeb-DF (CDF) \cite{li2020celeb}, DeepfakeDetection (DFD) \cite{roessler2019faceforensicspp}, DFDCP \cite{dolhansky2019deepfake}, FaceShifter (Fsh) \cite{roessler2019faceforensicspp}, UADFV \cite{yang2019exposing}, and DFDC \cite{dolhansky2020deepfake}. FF++ \cite{roessler2019faceforensicspp} is the most widely used dataset for evaluating the performance of deepfake detectors. It contains four manipulation methods: Deepfakes (DF)\cite{deepfakes}, FaceSwap (FS)\cite{faceswap}, Face2Face (F2F)\cite{thies2016face2face}, and NeuralTextures (NT)\cite{thies2019deferred}. FF++ has different compression levels: raw, c23 (high quality, HQ), and c40 (low quality, LQ). By default, the HQ version is used, and any deviation from this default is explicitly stated. To evaluate the adaptability of our framework, we perform experiments on six other widely used face-manipulated datasets.

\noindent \textbf{Evaluation Metrics.} We report the area under the curve (AUC) metric, where a higher value indicates better performance. In addition, we present the average AUC across all results and the number of times each method ranks in the top 3 across all test datasets for each evaluation.

\begin{table*}[h]
\centering
\caption{Cross-dataset AUC comparisons with top methods trained on FF++c23, marking best results in bold. }
\label{tab:cross-test}
\resizebox{\linewidth}{!}{%
\begin{tabular}{cccccccccc}
\toprule
Detector                         & CDFv1           & CDFv2           & DFD            & DFDC           & DFDCP          & Fsh             & UADFV           & Avg.                        & Top3       \\ \midrule
\multicolumn{1}{c|}{Meso4 \cite{afchar2018mesonet}}       & 0.7358          & 0.6091          & 0.5481         & 0.5560          & 0.5994         & 0.5660           & 0.7150           & \multicolumn{1}{c|}{0.6551} & 1          \\
\multicolumn{1}{c|}{MesoIncepion4 \cite{afchar2018mesonet}}   & 0.7366          & 0.6966          & 0.6069         & 0.6226         & 0.7561         & 0.6438          & 0.9049          & \multicolumn{1}{c|}{0.7364} & 3          \\
\multicolumn{1}{c|}{FWA \cite{li2018exposing}}         & 0.7897          & 0.6680           & 0.7403         & 0.6132         & 0.6375         & 0.5551          & 0.8539          & \multicolumn{1}{c|}{0.7239} & 1          \\
\multicolumn{1}{c|}{EfficientB4 \cite{tan2019efficientnet}} & 0.7909          & 0.7487          & 0.8148         & 0.6955         & 0.7283         & 0.6162          & 0.9472          & \multicolumn{1}{c|}{0.7718} & 3          \\
\multicolumn{1}{c|}{Capsule \cite{nguyen2019capsule}}     & 0.7909          & 0.7472          & 0.6841         & 0.6465         & 0.6568         & 0.6465          & 0.9078          & \multicolumn{1}{c|}{0.7488} & 2          \\
\multicolumn{1}{c|}{Xception \cite{chollet2017xception}}    & 0.7794          & 0.7365          & 0.8163         & 0.7077         & 0.7374         & 0.6249          & 0.9379          & \multicolumn{1}{c|}{0.7718} & 2          \\
\multicolumn{1}{c|}{CNN-Aug \cite{wang2020cnn}}     & 0.7420           & 0.7027          & 0.6464         & 0.6361         & 0.6170          & 0.5985          & 0.8739          & \multicolumn{1}{c|}{0.7020}  & 0          \\
\multicolumn{1}{c|}{F3Net \cite{qian2020thinking}}       & 0.7769          & 0.7352          & 0.7975         & 0.7021         & 0.7354         & 0.5914          & 0.9347          & \multicolumn{1}{c|}{0.7645} & 0          \\
\multicolumn{1}{c|}{FFD \cite{dang2020detection}}         & 0.7840           & 0.7435          & 0.8024         & 0.7029         & 0.7426         & 0.6056          & 0.9450           & \multicolumn{1}{c|}{0.7733} & 1          \\
\multicolumn{1}{c|}{Face X-ray \cite{li2020face}}  & 0.7093          & 0.6786          & 0.7655         & 0.6326         & 0.6942         & 0.6553          & 0.8989          & \multicolumn{1}{c|}{0.6985} & 0          \\
\multicolumn{1}{c|}{SPSL \cite{liu2021spatial}}        & 0.8150           & 0.7650           & 0.8122         & 0.7040          & 0.7408         & 0.6437          & 0.9424          & \multicolumn{1}{c|}{0.7875} & 3          \\
\multicolumn{1}{c|}{SRM \cite{luo2021generalizing}}         & 0.7926          & 0.7552          & 0.8120          & 0.6995         & 0.7408         & 0.6014          & 0.9427          & \multicolumn{1}{c|}{0.7760}  & 2          \\ 
\multicolumn{1}{c|}{CORE \cite{ni2022core}}        & 0.7798          & 0.7428          & 0.8018         & 0.7049         & 0.7341         & 0.6032          & 0.9412          & \multicolumn{1}{c|}{0.7694} & 0          \\
\multicolumn{1}{c|}{Recce \cite{cao2022end}}       & 0.7677          & 0.7319          & 0.8119         & 0.7133         & 0.7419         & 0.6095          & 0.9446          & \multicolumn{1}{c|}{0.7649} & 2          \\
\multicolumn{1}{c|}{UCF \cite{yan2023ucf}}         & 0.7793          & 0.7527          & 0.8074         & 0.7191         & 0.7594         & 0.6462          & 0.9528          & \multicolumn{1}{c|}{0.7801} & 5          \\ \midrule
Ours                             & \textbf{0.8693} & \textbf{0.7925} & \textbf{0.8749} & \textbf{0.7471} & \textbf{0.7790} & \textbf{0.6944} & \textbf{0.9551} & \textbf{0.8160}             & \textbf{8} \\ \bottomrule
\end{tabular}%
}
\end{table*}

\subsection{Intra-dataset Evaluation}
In this section, we perform a comprehensive evaluation of our proposed method against current state-of-the-art works on FF++c23 and FF++c40, as detailed in Table \ref{intra-test}. Specifically, our framework is trained on the FF++c23 and then tested on the full test sets of both the c23 and c40 versions. Furthermore, we perform separate tests on each manipulation method within the c23 version to thoroughly evaluate the performance of our method. The table shows the intra-test results of our method, highlighting its superior performance on the FF++c23 dataset with an outstanding AUC score of 0.9889, outperforming other approaches by 1.84\%. This top result, along with our competitive performance on the c40 dataset, contributes to the highest average AUC score of 0.9583 among all evaluated state-of-the-art deepfake detection methods, showing an improvement of 0.56\%. Notably, our approach consistently ranks in the top three in all categories, demonstrating its effectiveness and robustness in deepfake detection. These results underscore our method's leading edge in identifying deepfakes. 

\begin{table}[t]
\centering
\caption{Effectiveness of characteristic-conditioned knowledge, trained on FF++c23 and measured in AUC, with best results in bold.}
\label{ablation}
\resizebox{\linewidth}{!}{%
\begin{tabular}{ccccc}
\toprule
\multicolumn{2}{c|}{characteristic-conditioned}                             & \multirow{2}{*}{FF++ c23}        & \multirow{2}{*}{CDF v2}          & \multirow{2}{*}{DFDC}           \\ \cline{1-2}
specific                  & \multicolumn{1}{c|}{agnostic}                  &                                  &                                  &                                 \\ \midrule
                          & \multicolumn{1}{c|}{}                          & 0.9637                           & 0.7365                           & 0.7077                          \\
                          & \multicolumn{1}{c|}{\checkmark} & 0.9779                           & 0.7881                           & 0.7405                               \\
\checkmark & \multicolumn{1}{c|}{}                          & 0.9795                           & 0.7913                           & 0.7255                               \\
\checkmark & \multicolumn{1}{c|}{\checkmark} & \textbf{0.9889} & \textbf{0.7925} & \textbf{0.7471} \\
\bottomrule
\end{tabular}%
}
\end{table}

\subsection{Cross-dataset Evaluation}
In this section, we extend our evaluation to a cross-dataset test, assessing the robustness of our proposed method when applied to datasets beyond the training domain. Our framework, trained exclusively on the FF++c23 dataset, is subjected to a series of tests across multiple additional datasets, with the results compiled in Table \ref{tab:cross-test}. This rigorous testing protocol allows us to explore the generalization capabilities of our method. Despite the training being confined to FF++c23, our method demonstrates a consistent ability to detect deepfakes with high accuracy, as reflected by the impressive AUC scores across various datasets. This table delineates the cross-test results of our method, which again demonstrates its exceptional performance, this time across various datasets. Our approach outperforms other methods in terms of AUC score by 5.43\% on the CDFv1 dataset and 2.75\% on CDFv2, and achieves a notable improvement over existing methods across other challenging datasets, such as  2.80\% on DFDC, 1.96\% on DFDCP, 3.91\% on Fsh and a remarkable 5.86\% on DFD. The average AUC score for our method stands at 0.8160 with 2.85\% improvement compared to other methods, underscoring its robustness and consistency in performance across diverse testing scenarios. Furthermore, our method distinctly stands out by ranking in the top three positions all the times, more than any other method assessed, highlighting its leading capability in the field of deepfake detection across different datasets.

\subsection{Ablation Study}
We examine the characteristic-conditioned branches of our method and report the results in Table \ref{ablation}. 
In our ablation study, we evaluate the individual contributions of the characteristic-specific and characteristic-agnostic branches on the FF++c23, CDFv2, and DFDC datasets. Activating the characteristic-agnostic branch alone results in AUC increases of 1.42\%, 5.16\%, and 3.28\% across these datasets, respectively. The characteristic-specific branch alone yields a rise of 0.16\% and 0.32\% on FF++c23 and CDFv2, respectively, but experiences a slight decrease of 1.5\% on DFDC. The combined operation of both branches leads to the most substantial improvements, boosting AUC by 0.94\% on FF++c23, 0.12\% on CDFv2, and 0.66\% on DFDC, confirming the synergistic benefit of integrating both branches for more effective deepfake detection.

\section{Conclusion}
In this paper, we introduce AdaForensics, a characteristic-aware adaptive network for deepfake detection. our method leverages characteristic-conditioned knowledge and dynamically adjusts parameters to effectively handle diverse range of forgery artifacts of deepfakes. Our two-branch HyperNetwork architecture provides shareable abstractions and individual-level insights that are integrated into the target layer of the primary network. AdaForensics achieves state-of-the-art results in both intra-dataset and cross-dataset evaluations, demonstrating the effectiveness of our characteristic-conditioned knowledge in deepfake detection.

\section*{Acknowledgment}
This work was supported in part by the National Natural Science Foundation of China under Grant 62206147. This work is also supported in part by National Key R\&D Program of China under Grant 2021YFA0715202, Shenzhen Key Laboratory of Ubiquitous Data Enabling under Grant ZDSYS20220527171406015 and  the Shenzhen Science and Technology Program under Grant KQTD20170810150821146.

\bibliographystyle{IEEEbib}
\bibliography{reference}

\end{document}